\pdfoutput=1

\documentclass[11pt]{article}

\usepackage[final]{acl}

\usepackage{times}
\usepackage{latexsym}

\usepackage[T1]{fontenc}

\usepackage[utf8]{inputenc}

\usepackage{microtype}

\usepackage{inconsolata}

\usepackage{graphicx}

\usepackage{float}
\usepackage{multicol}
\usepackage{microtype}
\usepackage{graphicx}
\usepackage{booktabs}
\usepackage{amssymb}
\usepackage{amsmath}
\usepackage{multirow}
\usepackage{booktabs}
\usepackage{comment}
\usepackage{tikz}
\usepackage{listings}
\usepackage{xcolor}
\usepackage{array}
\usepackage{makecell}
\usepackage{lipsum}
\usepackage{cuted}
\usepackage{dirtytalk}
\usepackage{makecell}

\usepackage{pifont}
\newcommand{\cmark}{\ding{51}}%
\newcommand{\xmark}{\ding{55}}%

\usepackage{siunitx}
\definecolor{listingbg}{rgb}{0.97,0.97,0.97}
\definecolor{listingexample}{rgb}{0.55,0.7,0.9}
\definecolor{codegreen}{rgb}{0,0.6,0}
\definecolor{codegray}{rgb}{0.5,0.5,0.5}
\definecolor{codepurple}{rgb}{0.58,0,0.82}
\lstdefinestyle{mystyle}{
    backgroundcolor=\color{listingbg},
    commentstyle=\color{codegreen},
    keywordstyle=\color{magenta},
    numberstyle=\tiny\color{codegray},
    stringstyle=\color{codepurple},
    basicstyle=\fontsize{8pt}{9pt}\sffamily,
    breakatwhitespace=false,
    breaklines=true,
    captionpos=b,
    keepspaces=true,
    showspaces=false,
    showstringspaces=false,
    showtabs=false,
    tabsize=2,
    breakindent=0pt,
    frame=lines,
    aboveskip=\baselineskip,
    framesep=10pt,
    captionpos=b,
    abovecaptionskip=10pt,
    moredelim=**[is][\color{listingexample}]{@}{@},
    columns=fullflexible,
    numbers=none,
    xleftmargin=2pt,
    framexleftmargin=8pt,
}
\title{TSVer: A Benchmark for Fact Verification Against Time-Series Evidence}

\author{Marek Strong \and Andreas Vlachos \\
    Department of Computer Science and Technology \\
    University of Cambridge \\
    \{ms2518,av308\}@cam.ac.uk}

\begin{document}
\maketitle

\begin{abstract}

Reasoning over temporal and numerical data, such as time series, is a crucial aspect of fact-checking. While many systems have recently been developed to handle this form of evidence, their evaluation remains limited by existing datasets, which often lack structured evidence, provide insufficient justifications for verdicts, or rely on synthetic claims. In this paper, we introduce \textsc{TSVer}, a new benchmark dataset for fact verification focusing on temporal and numerical reasoning with time-series evidence. \textsc{TSVer} contains \num{304} real-world claims sourced from \num{41} fact-checking organizations and a curated database of \num{400} time series covering diverse domains.
Each claim is annotated with time frames across all pertinent time series, along with a verdict and justifications reflecting how the evidence is used to reach the verdict. Using an LLM-assisted multi-step annotation process, we improve the quality of our annotations and achieve an inter-annotator agreement of $\kappa = 0.77$ on verdicts. We also develop a baseline for verifying claims against time-series evidence and show that even the state-of-the-art reasoning models like \emph{Gemini-2.5-Pro} are challenged by time series, achieving a $63.57$ accuracy score on verdicts and an $\mathrm{Ev}^{2}\mathrm{R}$ score of $47.36$ on verdict justifications.

\end{abstract}
\section{Introduction}
\label{sec:introduction}

\renewcommand{\topfraction}{0.95}
\renewcommand{\textfraction}{0.05}
\renewcommand{\floatpagefraction}{0.9}

\begin{figure}
\centering
\includegraphics[width=1.0\linewidth,clip]{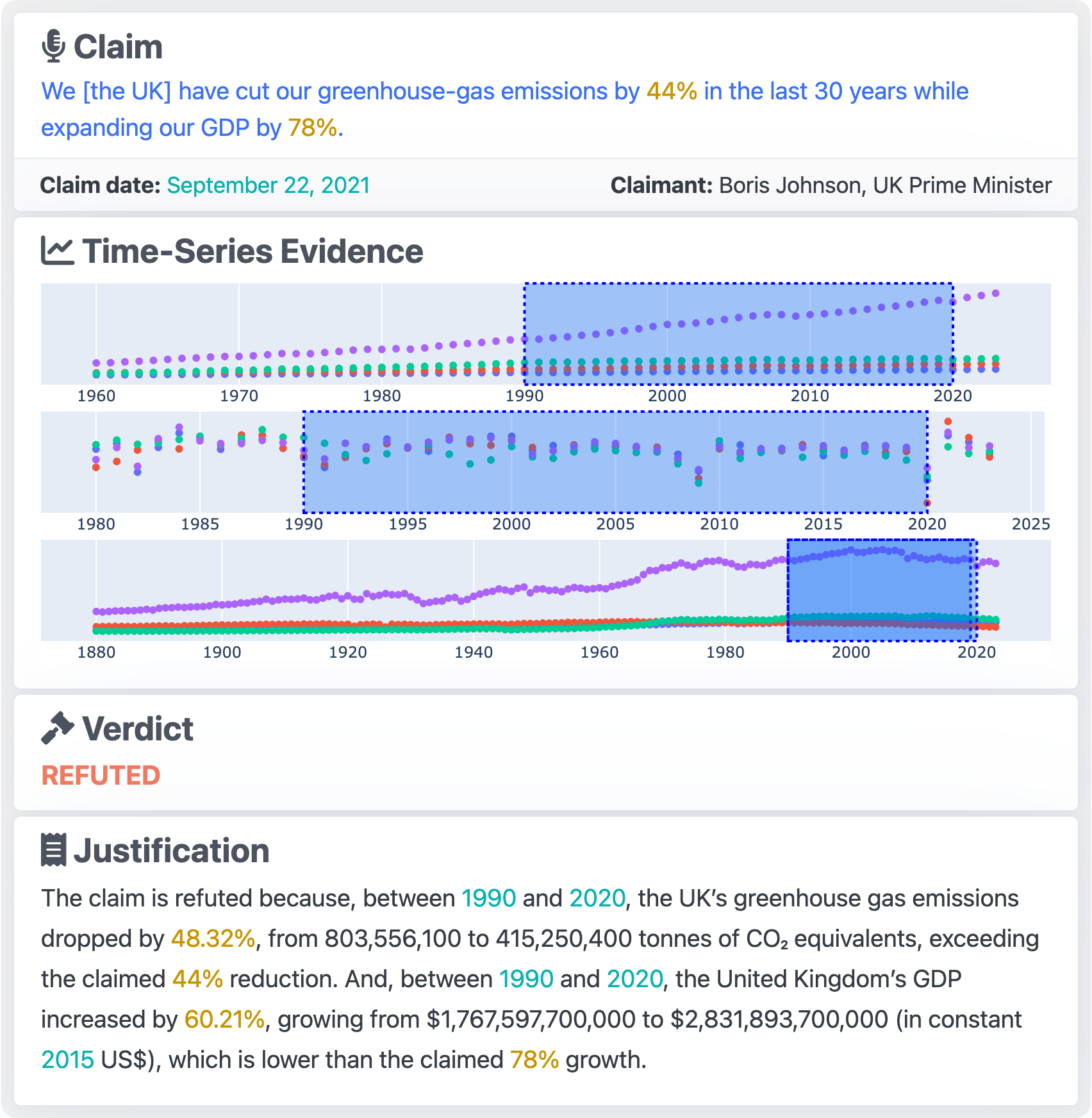}
\caption{\textbf{Example claim from TSVer.} Our dataset includes real-world claims paired with historical time-series evidence. All claims are annotated with time ranges (blue boxes), verdicts, and justifications emphasizing numerical and temporal reasoning.}
\label{fig:claimexample}
\end{figure}

With the growing use of social media and generative AI, there has been an unprecedented increase in the amount of inaccurate and misleading information \cite{adams2023why-370,arnold2020challenges}. In response, automated fact-checking systems 
have advanced substantially with the application of new large language models (LLMs) and the development of comprehensive datasets \cite{vykopal2024generative-316}. These systems have shown promising results in identifying and verifying claims across diverse domains and languages \cite{strong-etal-2024-zero}. However, they continue to face challenges when assessing claims that rely on external evidence \cite{fontana2025evaluating-be9} or when evaluating claims requires deeper reasoning beyond surface-level textual cues \cite{choi2023automated-51b,dziri2023faith-588}.

One of the areas where the reasoning limitations of LLMs are particularly prominent is numerical and temporal reasoning \cite{akhtar2023exploring-7c0,bubeck2023sparks-409}. This is particularly problematic in the context of fact-checking, where numerical and temporal expressions are prevalent. Studies have shown that over one-third of check-worthy claims involve numerical data \cite{v2024quantemp-0cb,hassan2017toward}, and many claims demand complex numerical reasoning to be properly evaluated \cite{aly2021feverous-c49}. Furthermore, as facts can evolve and change over time, fact-checking systems must be capable of accurately interpreting and reasoning over temporal aspects of both claims and supporting evidence \cite{allein2023implicit-16e,allein2020time-aware-a2c}. Therefore, it is increasingly important to effectively evaluate these capabilities and ensure that fact-checking systems can reliably reason over numerical and temporal data.


In this work, we introduce a new benchmark for evaluating fact verification systems using time-series evidence. Time series is a modality shown to be challenging for language models \citep{merrill2024language-854,fons2024evaluating-ff6} and is frequently used by human fact-checkers for fact verification \cite{akhtar2023chartcheck-7c4,alam2021survey-c90}. To address this gap, we present \textbf{\textsc{TSVer}}\textemdash the first benchmark dataset for explainable fact verification grounded in time-series evidence. \textsc{TSVer} pairs real-world claims with historical time-series evidence sourced from fact-checking organizations and includes textual justifications for verdicts, allowing for the evaluation of reasoning about evidence.

To construct \textsc{TSVer}, we collected \num{304} claims from \num{41 } fact-checking organizations, focusing on those involving numerical and temporal expressions resolvable via time series data. These claims were then aligned with our curated database of \num{400} time series, extracted from Our World in Data\footnote{\url{https://ourworldindata.org/}}. We avoided claims solvable by simple look-ups or simple arithmetic operations (common in numerical datasets \cite{lu2023scitab-f0e}) and instead targeted claims requiring reasoning across multiple countries, time series, and claims containing temporal and numerical ambiguities. While time series can be seen as tabular data, their temporal structure and scale add complexity.
Compared to prior datasets, \textsc{TSVer} features much larger time series, averaging around \num{11000} records per instance, with some over \num{80000} records, posing new challenges for fact verification on high-volume, real-world data.

Figure \ref{fig:claimexample} illustrates an example claim from TSVer with annotations for the evidence, verdict, and justification. To fact-check this claim, a system must identify the relevant time series from our dataset (i.e., Greenhouse Gas Emissions, Gross Domestic Product, and Annual GDP Growth), determine the relevant time frames (i.e., 1990–2020), reason over all data points within these time frames for relevant countries (i.e., the United Kingdom), and generate a verdict accompanied by a justification. Identifying relevant time frames is particularly challenging in our benchmark, as selecting different date ranges often leads to different verdicts. Politicians frequently exploit this by choosing selective dates to support their claims, a practice known as \textit{cherry-picking} \cite{asudeh2020detecting-3ea}. Additionally, a time series may contain both supporting and contradicting periods.
For example, while the Gemini-2.5 Pro reasoning model correctly selects 1990 as the starting year (a common baseline for climate-related claims), it uses the period 1990–2019 rather than 1990–2020 to provide supporting evidence. This contradicts the reasoning of our annotators and the original fact-checking article, which notes that Boris Johnson, who made the claim in 2021, relied on outdated figures that ignore the extraordinary impact of the COVID-19 pandemic.

We also propose a fact-checking pipeline as a baseline to demonstrate the feasibility of the task and to benchmark the performance of the state-of-the-art open-weight and proprietary language models. The \textit{gemini-2.5-pro} reasoning model \cite{team2023gemini}, currently ranked first on ChatBot Arena \cite{chiang2024chatbot-e45}, achieves an accuracy of \num{63.57} on verdict prediction. Additionally, to evaluate models' reasoning in comparison to human annotators, we use the $\mathrm{Ev}^{2}\mathrm{R}$ scorer \cite{akhtar2024ev2r-3f6}, originally developed for evidence retrieval, and demonstrate its effectiveness in this new context. Furthermore, to specifically evaluate evidence retrieval performance with time series data, we introduce a novel metric\textemdash \textsc{TSCS}, which jointly measures the accuracy of both time series selection and temporal coverage.


Our dataset is available under a \textsc{CC BY-SA 4.0} license at \url{https://github.com/marekstrong/TSVer}.


\section{Related Work}
\label{sec:literature_review}

We summarize key characteristics of existing fact verification datasets in Table~\ref{tab:litreview}. In the following , we compare these datasets to \textsc{TSVer} along three key dimensions: evidence modalities, numerical and temporal focus, and the inclusion of human-written justifications.

\begin{table*}[ht!]
    \centering
    \resizebox{1.0\linewidth}{!}{

\begin{tabular}{@{}l|lcccccc@{}}
\toprule
\textbf{Dataset} & \textbf{Domain} &\textbf{\#Labels}&\makecell{\textbf{Real-world}\\\textbf{Claims}} & \makecell{\textbf{Numerical}\\\textbf{Focus}}& \makecell{\textbf{Temporal}\\\textbf{Focus}} & \makecell{\textbf{Evidence}\\\textbf{Modality}} & \textbf{Justifications} \\
\midrule
 FEVER \cite{thorne-etal-2018-fever}&   Multi&3&\xmark& \xmark& \xmark& Text& \xmark\\
TabFact \cite{chen2019tabfact-c37}          &                              Multi&2&\xmark&                          \cmark&                         \xmark&                            Tables&                         \xmark \\
FEVEROUS \cite{aly2021feverous-c49}         &                              Multi&3&\xmark&                          \xmark&                         \xmark&                            Text + Tables&                         \xmark \\
AVERITEC \cite{schlichtkrull2023averitec-433}         &                              Multi&4&\cmark&                          \xmark&                         \xmark&                            Text&                         \cmark \\
SciTab \cite{lu2023scitab-f0e}           &                              Science&3&\cmark&                          \cmark&                         \xmark&                            Tables&                         \xmark \\
 Liar++ \cite{russo2023benchmarking-1a4} &  Politics&2& \cmark& \xmark& \xmark& Text&\cmark\\
 T-FEVER \cite{barik2024time-3ea}& Multi& 3& \xmark& \xmark& \cmark& Text&\xmark\\
 T-FEVEROUS \cite{barik2024time-3ea}       &                              Multi&3&\xmark&                          \xmark&                         \cmark&                            Text + Tables&                         \xmark \\
 ChronoClaims \cite{barik2024chronofact-e61}     &                              Multi&3&\xmark&                          \xmark&                         \cmark&                            Text&                         \xmark \\
QuanTemp \cite{v2024quantemp-0cb}         &                              Multi&3&\cmark&                          \cmark&                         \cmark&                            Text&                         \cmark\\
 FinDVer \cite{zhao2024findver-7a4}&  Finance&2& \xmark& \cmark& \xmark& Text + Tables&\cmark\\
 \midrule
\textbf{TSVer}            &                              Multi&4&\cmark&                          \cmark&                         \cmark&                            Time Series&                         \cmark \\ \bottomrule
\end{tabular}

    }
    \caption{Comparison of TSVer with other fact-checking datasets.}
    \label{tab:litreview}
\end{table*}

\paragraph{Evidence Modalities}

Early fact-checking datasets, such as FEVER \cite{thorne-etal-2018-fever}, primarily relied on textual evidence to support or refute claims. However, as a substantial portion of factual information is embedded in structured sources (e.g., tables, knowledge bases, time series), subsequent datasets have expanded to include these modalities as well. FEVEROUS \cite{aly2021feverous-c49} extends the FEVER framework by pairing synthetic claims with both textual and tabular evidence. FinDVer \cite{zhao2024findver-7a4} focuses on tabular data extracted from financial reports, linking it to relevant claims. SciTab \cite{lu2023scitab-f0e} compiles real-world claims from scientific literature and supports them with tabular evidence. Compared to the previous datasets, \textsc{TSVer} introduces time series as the primary source of evidence.


\paragraph{Numerical and Temporal Focus}

Since numerical and temporal expressions are common in fact-checking, recent datasets increasingly focus on these aspects. Several benchmarks target numerical reasoning with structured data: TabFact \cite{chen2019tabfact-c37} verifies crowd-sourced claims against Wikipedia tables, while SciTab \cite{lu2023scitab-f0e} uses scientific tables to assess compositional reasoning. Domain-specific datasets like FinDVer (finance) \cite{zhao2024findver-7a4} combine text and tables with an emphasis on numerical calculations. In open-domain fact-checking, QuanTemp \cite{v2024quantemp-0cb} introduces real-world claims involving numerical comparisons and trends, explicitly incorporating temporal reasoning. It is the only existing dataset that targets both numerical and temporal expressions, and since its claims are also sourced from fact-checking websites, it is the closest to our work.
While QuanTemp includes justifications, they are unstructured and exclusively textual. In contrast, \textsc{TSVer} provides high-quality, structured justifications for time-series data, including annotations that explain how specific time frames support a claim. Moreover, time series are central in \textsc{TSVer}, whereas it is only one of several claim categories in QuanTemp. Other temporal datasets such as T-FEVER / T-FEVEROUS \cite{barik2024time-3ea}, and ChronoClaims \cite{barik2024chronofact-e61} focus more narrowly on date-sensitive or chronological assertions, often using synthetic augmentation or curated timelines.





\paragraph{Justifications}

Justifying claim verification decisions is a critical component of journalistic fact-checking, reflecting the broader need for transparency and accountability in the verification process \citep{guo2022survey-a32,kotonya2020explainable-765}. \citet{warren2025show-4d4} recently argued that the inherent complexity of fact-checking requires automated systems to offer justifications that allow fact-checkers to critically evaluate their results. Unfortunately, most of the aforementioned datasets do not provide such rationale. A few recent datasets aim to address this gap: AVeriTeC \cite{schlichtkrull2023averitec-433} adds textual explanations synthesizing evidence, LIAR++ \cite{russo2023benchmarking-1a4} includes journalist-written justifications, and FinDVer \cite{zhao2024findver-7a4} provides expert-annotated step-by-step reasoning.

\paragraph{}
To the best of our knowledge, \textsc{TSVer} is the only dataset that provides complex structured evidence, focuses on both numerical and temporal claims, and provides justifications for claims' verdicts as well as retrieved evidence. A detailed comparison of \textsc{TSVer} with existing fact-checking datasets across these dimensions is presented in Table~\ref{tab:litreview}.








\section{Annotation Process}

\begin{figure*}[th!]
\centering
\includegraphics[width=1.0\linewidth, trim={0.5cm 0.1cm 0.5cm 0.1cm},clip]{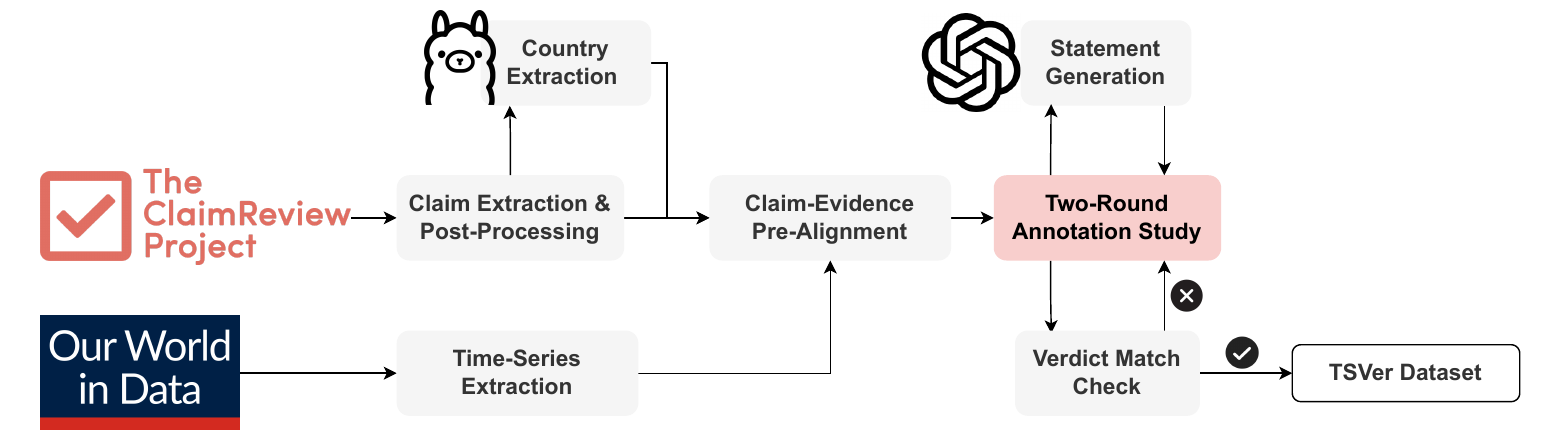}
\caption{\textbf{TSVer data collection pipeline.}}
\label{fig:overallpipeline}
\end{figure*}

This chapter describes the construction of the \textsc{TSVer} dataset, detailing the end-to-end pipeline from claim extraction to evidence alignment and annotation. An overview of the entire process is illustrated in Figure~\ref{fig:overallpipeline}.

\subsection{Claim Extraction and Post-Processing}
\label{sec:claim_postprocessing}

To construct our dataset, we began by collecting an initial pool of around six thousand claims using The Google FactCheck Claim Search API\footnote{\url{https://toolbox.google.com/factcheck/apis}}, which aggregates content from the ClaimReview project\footnote{\url{https://www.claimreviewproject.com/}}. We then applied a filtering pipeline to identify claims likely to require reasoning over temporal and numerical data. Specifically, we used \textit{HeidelTime} \cite{strotgen-gertz-2010-heideltime} to detect the presence of temporal expressions and \textit{spaCy} \cite{Honnibal_spaCy_Industrial-strength_Natural_2020} to extract numerical expressions from the claim texts. Each retained claim was linked to its corresponding fact-checking article and associated metadata, including the publisher, claimant, and claim date, as provided by the API.

We then post-processed the extracted claims in three steps. First, we manually reviewed cases with missing claim dates and inferred the dates from the accompanying articles. Second, for claims that did not mention any country, we added annotations in square brackets where appropriate (e.g., the annotation '[the UK]' as shown in Figure~\ref{fig:claimexample}). Lastly, we provided an additional annotation for all countries mentioned in the corresponding fact-checking article. For this step, we prompted the \textit{\mbox{Llama-3.1-8B}} model \cite{dubey2024llama-906} to identify country names within the raw HTML of each article.


\subsection{Time-Series Extraction}

As the source of our time series data, we selected Our World in Data (OWID)\footnote{\url{https://ourworldindata.org/}}, a non-profit online publication that compiles and curates open-access datasets on global issues such as population, health, economic development, environment, and governance. OWID’s focus on socially and politically relevant topics aligns well with the kinds of subjects that are frequently addressed in professional fact-checking. Moreover, OWID is a useful source as it integrates data from a variety of original sources and applies consistent post-processing steps, including the standardization of country names and regional groupings, unit normalization, and computation of derived indicators (e.g., per capita values).

To constrain the complexity of the dataset and simplify alignment with claims, we restricted our scope to time series reported at an annual resolution. However, we also included data sources for which meaningful annual aggregates could be computed from higher-frequency data. In total, we selected \num{400} time series spanning a wide range of domains.
Each time series was paired with metadata provided by OWID, including its title, description (including associated notes), and units. Examples of this metadata can be found in Appendix~\ref{sec:appendix_datasetdetails}.

\subsection{Claim-Evidence Pre-Alignment}

Aligning collected claims with relevant time-series evidence presents some major challenges; in many cases, identifying the correct time series can be as difficult as verifying the claim itself.
Time-series data often include closely related but distinct indicators (e.g., "Cumulative CO$_{2}$ Emissions", "Per-capita CO$_{2}$ Emissions", and "CO$_{2}$ Emissions"), and for some indicators, data may be available from multiple sources (e.g., "OECD Tax Revenue as Share of GDP" vs. "UNU-WIDER Tax Revenue as Share of GDP").
Furthermore, most claims require alignment with multiple time series to be adequately verified (see Figure~\ref{fig:claimexample}).

Given this complexity, we opted to leave the final alignment of claims with evidence to human annotators (see Section~\ref{sec:annotation-study}). However, to reduce annotation effort and ensure consistency, we introduced a pre-alignment step. Specifically, we grouped time series into semantic categories using the OWID taxonomy and matched claims to these groups using keyword-based heuristics. For example, claims containing the term "emissions" were pre-aligned with all time series in the "Environment" category. Claims that could not be aligned with any group were excluded from the dataset.


\subsection{Two-Round Annotation Study}
\label{sec:annotation-study}

We conducted a two-round annotation study using the Prolific platform\footnote{\url{https://www.prolific.com/}}. Examples of annotation interfaces are shown in Appendix \ref{sec:appendix_interface}.

In \textbf{Phase 1}, annotators were given a claim, its corresponding fact-checking article, and a set of potentially relevant time series. Their task was to: (i)~select relevant time series, (ii)~identify all time ranges useful for verification, and (iii)~provide explanations for each selected range. Since annotators had access to full articles, they could follow the reasoning of professional fact-checkers, which often included contextual knowledge. For example, fact-checkers often reference years 2005 (Kyoto Protocol) or 2016 (Paris Agreement) in climate-related claims. However, as fact-checkers may have relied on sources beyond OWID, we instructed annotators to prioritize only the provided time-series evidence. To capture contextual information, annotators were also asked to record any useful details from the fact-checking articles (e.g., for resolving ambiguities) that informed their choices of time series and time ranges. These notes will be released as part of the dataset.

In \textbf{Phase 2}, new annotators reviewed the evidence annotated in Phase 1 without access to the articles. They were asked to assign one of four verdict labels (see Section~\ref{sec:dataset_stats}) based on the claim and evidence, and to provide a justification. To assist with reasoning, we presented precomputed statistics (e.g., min/max values, averages, trends) for each time range. The full list is in Appendix~\ref{sec:appendix_annotation_details}. 
To avoid overwhelming annotators with irrelevant data, we limited the statistics shown to countries mentioned in the article, as identified during the LLM-based post-processing step (see Section~\ref{sec:claim_postprocessing}).


Our initial annotation results showed that while these statistics helped, annotator justifications often lacked numerical expressions and reasoning. To address this, we prompted \textit{gpt-4o-2024-11-20} \cite{openai2023gpt-4-0bf} to generate up to five statements based on the provided time-series ranges, focusing on numerical expressions (see Appendix \ref{sec:appendix-prompting} for prompting details). Annotators were asked to identify truthful ones and optionally incorporate them into their justifications. This modification substantially increased the use of numerical details, resulting in justifications that were more precise and better aligned with the presented time-series data.

As a quality check, we compared annotator verdicts to those from the reference fact-checks. If a majority verdict disagreed with the reference verdict from the article, the claim was re-annotated in a second round. However, we kept the claims afterwards as differences in the evidence may naturally lead to differing verdicts.

\subsection{Inter-Annotator Agreement}

Following \citet{schlichtkrull2023averitec-433} and \citet{ousidhoum-etal-2022-varifocal}, we measured inter-annotator agreement using Randolph’s free marginal multi-rater $\kappa$ \cite{randolph2005free}. For verdict labels, we achieve an agreement score of $\kappa = 0.77$, indicating substantial agreement among annotators. For the selection of numerical statements generated by GPT-4o, we observe a lower agreement of $\kappa = 0.61$.

\section{\textsc{TSVer} Benchmark}

\subsection{Dataset Statistics}
\label{sec:dataset_stats}

We collected \num{304} claims from a total of \num{41} fact-checking sites. The most represented sites were \textit{Africa Check (21.38\%)}, \textit{Full Fact (15.13\%)}, and \textit{PolitiFact (12.50\%)} (see Appendix~\ref{sec:appendix_datasetdetails} for the full distribution by organization). Following the approach of \citet{schlichtkrull2023averitec-433}, we adopt a \num{4}-class labeling scheme: \textit{SUPPORTS}, \textit{REFUTES}, \textit{NOT ENOUGH INFO}, and \textit{CONFLICTING EVIDENCE/CHERRY-PICKING}. The dataset is inherently unbalanced, with a higher proportion of \textit{REFUTES} labels (53.95\%). This reflects the nature of fact-checking workflows, where journalists often prioritize addressing false or misleading claims. In terms of geographic coverage, the most claims discuss the United States (29.93\%), followed by the United Kingdom (25.99\%), Australia (17.11\%), India (16.12\%), and Nigeria (14.80\%). A more detailed country-level distribution is provided in Appendix~\ref{sec:appendix_datasetdetails}.

\textsc{TSVer} also includes a curated collection of \num{400} time series from \textit{Our World in Data}. All time series were preprocessed into a consistent format, and we provide both titles and descriptions to facilitate retrieval. Additionally, inspired by the time-series taxonomy introduced by \citet{fons2024evaluating-ff6}, we categorize each series by feature type (e.g., trend, volatility, stationarity). To this end, we prompted \textit{Gemini-2.5-Pro} to generate descriptive features for each time series and annotated time ranges.

Few-shot prompting, a technique in which a model is given a few in-context examples, has been shown to improve performance across various fact-checking tasks, including claim detection, evidence retrieval, and general reasoning \cite{vykopal2024generative-316,li2023self-checker-8ac,wang2023check-covid-884}. Thus, to support few-shot prompting for our benchmark, we set aside a small development set of \num{24} claims. To ensure minimal overlap with the test set, these claims are drawn from entirely separate domains, which are explicitly excluded from the test data.

\subsection{Synthetic Claims}

To further improve the practical utility of this benchmark for training and evaluation, we augmented \textsc{TSVer} by modifying the countries and dates mentioned in claims. We used \textit{\mbox{gemini-2.5-pro}} to guide this process, generating new claims with different labels. This approach resulted in \num{400} additional synthetic claims, which will be released as a separate dataset within \textsc{TSVer}. However, this synthetic dataset was not used in our main experiments reported in Section~\ref{sec:sec_experiments}.

\subsection{Baseline Pipeline}

Our baseline pipeline consists of two main components: (1) time-series retrieval and (2) verdict and justification generation.

To retrieve relevant time-series evidence, we rely on the textual metadata (e.g., title, description, units) associated with each time series in the database. Examples of this metadata are provided in Appendix \ref{sec:appendix_datasetdetails}. While directly using raw time-series data could provide better retrieval performance, exploring methods such as time series encoders \cite{woo2024unified-d51} is beyond the scope of this work.

We prompt an LLM in a few-shot setup, providing examples from the development set, to generate a list of relevant time series as evidence. However, this initial retrieval step often yields too much data: even a few complete time series can exceed the input limits of most LLMs. For example, using Gemini-2.5-Pro, only 31\% of the cases had retrieved evidence with fewer than 1 million tokens. Therefore, we apply additional filtering using the same LLM (in a few-shot set-up) to further refine the results. Specifically, we prompt the model to identify relevant time ranges and relevant countries. Note that the model does not have access to fact-checking articles during testing, so it cannot leverage country information in those articles as was done during the annotation phase.

The second baseline component starts by loading the specified slices of time-series data according to the retrieval results. We then prompt the same LLM to generate a verdict along with supporting justifications. Due to the large input size, we adopt a zero-shot setup for this stage. Additionally, for non-reasoning models, we apply Chain-of-Thought prompting \cite{wei2022chain-f8a} to explicitly encourage step-by-step reasoning in the output.

All prompting templates are reported in Appendix \ref{sec:appendix-prompting}.

\subsection{Baseline LLMs}

Due to the large input sizes resulting from representing time series data in raw text format, our evaluation is restricted to language models with extended context windows. Specifically, we consider only models that support a minimum of \num{128}k tokens. Among proprietary models, we include \textit{Gemini} \cite{team2023gemini} and \textit{GPT} \cite{openai2023gpt-4-0bf}, while for open-weight models, we choose \textit{Mistral} \cite{jiang2023mistral-6be} and \textit{Llama} \cite{dubey2024llama-906}.

For all experiments, the temperature is set to \num{0.01}, top-p to \num{0.95}, and the maximum output length is \num{4096} tokens.

\section{Experiments}
\label{sec:sec_experiments}

\subsection{Evaluation Metrics}


In addition to standard verdict prediction metrics such as macro-F1 and accuracy, we introduce two complementary evaluation metrics to assess the effectiveness of our retrieval and justification components. These metrics specifically capture the accuracy of time series retrieval with temporal alignment and the factual consistency of generated justifications relative to human justifications.

\subsubsection{Time Series Coverage Score}

To evaluate the performance of the retrieval component, we assess two key aspects with respect to human-annotated ground truth: (1) whether the correct time series datasets are retrieved, and (2) how well the retrieved time ranges align with the annotated relevant time spans. Both over-retrieval (e.g., retrieving more datasets or time spans than necessary) and under-retrieval (e.g., omitting relevant time series or time spans) can negatively impact downstream performance, either by exceeding the context window or by failing to provide sufficient evidence for verification.

We propose the Time Series Coverage Score (TSCS), a metric that jointly captures the accuracy of both time series selection and temporal coverage. TSCS combines a dataset-level F1 score with a temporal Jaccard Index to evaluate the quality of each retrieval instance.

\begin{equation}
\text{TSCS} = \frac{1}{N} \sum_{i=1}^{N} \left( \text{F1}_i \cdot \overline{J}_i \right)
\label{eq:tscs}
\end{equation}

In Equation~\ref{eq:tscs}, $N$ denotes the number of evaluation instances. For each instance $i$, the F1 score is computed, reflecting whether the correct set of time series datasets was retrieved. $\overline{J}_i$ is then the average Jaccard Index over the matched datasets, measuring temporal alignment.

The average Jaccard Index is defined as:

\begin{equation}
\overline{J} = \frac{1}{T} \sum_{j=1}^{T} \frac{|\hat{Y}_j \cap Y_j|}{|\hat{Y}_j \cup Y_j|}
\label{eq:jaccard_general}
\end{equation}

Here, $T$ is the number of retrieved time series that correctly match the ground truth data. For each time series $j$, $\hat{Y}_j$ and $Y_j$ represent the annotated and retrieved time ranges, respectively. The Jaccard Index measures the degree of overlap between these ranges. Averaging across all matched time series yields a robust estimate of temporal accuracy.

\subsubsection{$\mathrm{Ev}^{2}\mathrm{R}$ Score for Justifications}

The $\mathrm{Ev}^{2}\mathrm{R}$ scorer \cite{akhtar2024ev2r-3f6} evaluates the quality of evidence retrieval in automated fact-checking by comparing retrieved evidence against reference evidence through atomic fact decomposition. Since the metric essentially compares two free-form texts for factual overlap, we test its suitability for comparing verdict justifications in this context.

The metric comprises three components:

\paragraph{Precision ($s_{\text{prec}}$)}
This measures the proportion of atomic facts in the retrieved evidence ($A_{\hat{E}}$) that are supported by the reference evidence ($E$). It is calculated as:

\newcommand{\Ind}[1]{\mathbb{I}\!\left[#1\right]}

\begin{equation}
s_{\text{prec}} \;=\;
\frac{1}{\lvert A_{\hat{E}}\rvert}
\sum_{a_{\hat{E}}\in A_{\hat{E}}}
      \Ind{\,a_{\hat{E}}\text{ supported by }E}
\label{eq:ev2r-precision}
\end{equation}

\paragraph{Recall ($s_{\text{recall}}$)}
This assesses the proportion of atomic facts in the reference evidence ($A_E$) that are supported by the retrieved evidence ($\hat{E}$). It is given by:

\begin{equation}
s_{\text{recall}} \;=\;
\frac{1}{\lvert A_{E}\rvert}
\sum_{a_E\in A_E}
      \Ind{\,a_{E}\text{ supported by }\hat{E}}
\label{eq:ev2r-recall}
\end{equation}

\paragraph{F1 Score ($s_{F_1}$)} This is the harmonic mean of precision and recall, providing a balanced measure of the retrieval quality:

\begin{equation}
s_{F_1} \;=\;
\frac{2\,s_{\text{prec}}\,s_{\text{recall}}}
     {s_{\text{prec}} + s_{\text{recall}}}
\label{eq:ev2r-f1}
\end{equation}

In this work, we use the reference-based atomic score from $\mathrm{Ev}^{2}\mathrm{R}$ \citep{akhtar2024ev2r-3f6}, which was inspired by FactScore \cite{min-etal-2023-factscore}. While we follow a similar prompting template, we adapt it using modified examples drawn from our development set as few-shot instances. We use \textit{\mbox{gemini-2.5-flash-preview-09-2025}} as the scorer model, and the prompting details can be seen in Appendix \ref{sec:appendix-prompting}.

\begin{table*}[]
    \centering
    \resizebox{1.0\linewidth}{!}{

\begin{tabular}{@{}lccclcc@{}lccc}
\toprule
                              &&& Time Series&  &\multicolumn{2}{c}{Verdicts}&  &\multicolumn{2}{c}{Justifications} &\\ \midrule
Model                         & Params&Max Tokens& TSCS &\hspace{0em}&F1            & Accuracy           & \hspace{1em}&METEOR          & $\mathrm{Ev}^{2}\mathrm{R}$ &CL Errors\\ \midrule
Gemini-2.5-pro-06-17&- &1M&                27.87&  &54.10& 63.57&  &30.26&47.36&2.86 \%\\
 GPT-5.2-2025-12-11&- &400k& 25.83& & 52.63& 62.86& & 26.14&46.51&4.64 \%\\
 Mistral-large-2512&675B &256k& 20.01& & 48.06& 62.14& & 32.13&40.85&5.00 \%\\
 Ministral-8b-2512& 8B &256k& 11.03& & 42.31& 54.29& & 30.16& 35.64&8.57 \%\\
 Ministral-3b-2512& 3B& 256k& 11.32& & 38.60& 49.29& & 29.04& 34.06&16.79 \%\\
 Llama-3.3-70B &70B &128k& 15.41& & 39.61& 55.36& & 31.36&43.98&8.21 \%\\
Llama-3.1-8B &8B &128k&                3.19&  &24.46& 23.21&  &27.64&25.07&42.50 \%\\ \bottomrule
\end{tabular}

    }
    \caption{Verification results with baseline models on the \textsc{TSVer} test set.}
    \label{tab:main-results}
\end{table*}

\subsection{Results}

Our main evaluation results are reported in Table~\ref{tab:main-results}.
We can observe that even state-of-the-art API models such as \textit{Gemini-2.5} and \textit{GPT-5.2} struggle with the \textsc{TSVer} benchmark, achieving verdict prediction accuracies of \num{63.57} and \num{62.86}, respectively. This indicates that a large portion of the benchmark remains challenging, even for the most capable commercial models. In contrast, smaller open-weight models, including \textit{Ministral-8B}, \textit{Ministral-3B}, and \textit{Llama-3.1-8B}, perform substantially worse, with accuracies of \num{54.29}, \num{49.29}, and just \num{23.21}, respectively. These results underscore both the difficulty of \textsc{TSVer} and its effectiveness as a probing tool for evaluating model reasoning and verification capabilities.


When evaluating retrieval quality using the Time Series Coverage Score (TSCS), we observe a substantial performance gap across models. \textit{Gemini} achieves the highest TSCS at \num{27.87}, followed by \textit{GPT} with a score of \num{25.83}, indicating that these models are more effective at both selecting the relevant time series and aligning the retrieved time ranges with the human-annotated spans. In contrast, smaller models such as \textit{Ministral-8B}, \textit{Ministral-3B}, and \textit{Llama-3.1-8B} perform considerably worse, with TSCS values of \num{11.03}, \num{11.32}, and just \num{3.19}, respectively.

Further analysis of low TSCS scores reveals that smaller models tend to over-retrieve. For instance, while \textit{Gemini} and \textit{GPT} retrieve \num{2728} and \num{2143} time series in total across the test set, \textit{Ministral-8B} and \textit{Llama-3.1-8B} retrieve substantially more\textemdash\num{17108} and \num{31419}, respectively. This excessive retrieval not only increases the difficulty of downstream reasoning tasks but also places greater demands on context length. As shown in Table~\ref{tab:main-results}, smaller models are more likely to exceed their context window limits, leading to context length (CL) inference errors. Notably, \textit{Llama-3.1-8B} failed on \num{42.5}\% of the test instances due to exceeding its \num{128}k token limit.


We evaluate justification quality using both \mbox{METEOR} \cite{banerjee-lavie-2005-meteor} and the $\mathrm{Ev}^{2}\mathrm{R}$ score. While METEOR provides a surface-level measure of lexical overlap with reference justifications, it shows a limited capability to differentiate between our models. For instance, \mbox{\textit{Llama-3.1-8B}} and \textit{GPT-5.2} obtain comparable \mbox{METEOR} scores of \num{27.64} and \num{26.14}, despite a substantial gap in their verdict and retrieval performance. This aligns with prior findings of \citet{akhtar2024ev2r-3f6}, which suggest that surface metrics like METEOR often fail to correlate with human judgments of factual adequacy in explanations.

In contrast, $\mathrm{Ev}^{2}\mathrm{R}$ provides a more informative signal by evaluating the factual alignment between model-generated and reference justifications via atomic fact decomposition. According to this metric, \textit{Gemini} leads with an $\mathrm{Ev}^{2}\mathrm{R}$ score of \num{47.36}, followed by \textit{GPT} at \num{46.51}. All smaller models score lower, with \textit{Ministral-8B}, \textit{Ministral-3B}, and \textit{Llama-3.1-8B} scoring only \num{35.64}, \num{34.06}, and \num{25.07}, respectively. These results suggest that $\mathrm{Ev}^{2}\mathrm{R}$ is more sensitive to factual accuracy and evidence relevance in generated justifications, making it a more reliable indicator of model capability in complex verification tasks.

To further probe the complexity of our benchmark, we also conducted experiments with PASTA \cite{gu-etal-2022-pasta}, an NLI model specifically designed and pre-trained for numerical and tabular reasoning. PASTA aligns well with \textsc{TSVer}, since it can reason over table-based operations such as column aggregation, min/max comparisons, and row filtering, operations that are commonly required in our dataset. Using the authors’ publicly released checkpoint\footnote{\url{https://github.com/ruc-datalab/PASTA}}, we fine-tuned the model on the TabFact dataset \cite{chen2019tabfact-c37} and applied PASTA's linearization scripts to convert our time-series tables into a format compatible with the model. Since PASTA only performs binary fact verification, we collapsed all labels other than \textit{SUPPORTED} into the \textit{REFUTED} category. Under this setup, PASTA achieved an F1 score of \num{43.56}, underscoring both the difficulty of our benchmark and the current limitations of table-aware NLI methods when applied to time-series reasoning.

\subsection{Discussion}




Our baseline system employs a straightforward strategy: formatting raw time series data as Markdown-style tables using \texttt{pandas}\footnote{\url{https://pandas.pydata.org/docs/reference/api/pandas.DataFrame.to_markdown.html}}. While this representation enables seamless integration with existing LLM-based systems, it presents challenges due to the input length and the continuous, numerical nature of time series data. In particular, the tokenization of floating-point numbers using byte pair encoding (BPE) can yield inconsistent and inefficient representations \cite{gruver2023large-ef1,spathis2023first-b19}. Furthermore, the sheer length of many time series, often spanning thousands or even millions of data points, can easily exceed the context window limitations of current LLMs.

Unlike traditional fact verification tasks, where input length can often be managed by selecting the top-N most relevant sentences, time series data does not lend itself to such straightforward truncation. Relevant temporal patterns may span long, continuous ranges, making it harder to reduce input size without losing critical evidence.

To address these issues, several studies have introduced quantization-based techniques. For example, models like SpeechGPT \cite{zhang2023speechgpt-89d} and AudioLM \cite{borsos2022audiolm-94f} employ K-means clustering to convert continuous signals into discrete token sequences, while others use VQ-VAE for a similar discretization process \cite{duan2023dewave-8e6,strong2021discrete-c06}. Alternative strategies involve integrating dedicated time series encoders with LLMs, as seen in models such as GPT4TS \cite{zhou2023one-139} and Time-LLM \cite{jin2023time-llm-41b}.


Since our results highlight retrieval quality as a critical bottleneck, particularly for smaller models, exploring more efficient time series representations may enable future systems to better encode and reason over temporal data within constrained model budgets.

\section{Conclusion}

We introduced \textsc{TSVer}, the first benchmark dataset for fact verification grounded in real-world time-series evidence. By focusing on complex claims requiring numerical and temporal reasoning, \textsc{TSVer} show the limitations of current fact-checking systems and large language models in handling structured temporal data. Our LLM-assisted annotation pipeline enables the alignment of claims with time-series evidence, achieving a substantial inter-annotator agreement of $\kappa = 0.77$ on verdicts. The dataset supports rigorous evaluation of both evidence selection and reasoning quality, and we hope \textsc{TSVer} will serve as a valuable resource for advancing research in explainable and evidence-based fact verification.

\section*{Limitations}

\paragraph{Lack of Multilingual Coverage}

Although our claims span topics and entities from many parts of the world, we only collected claims and fact-checking articles in English. This design choice simplifies annotation and model evaluation, yet it also means that the benchmark does not assess cross-lingual retrieval, multilingual reasoning, or language-specific numeral and date formats.

\paragraph{Source Bias and Coverage Limitations}

The claims in \textsc{TSVer} are sourced directly from existing fact-checking articles via the Google Fact Check Explorer. As such, our dataset inherits any biases or limitations present in those original articles. Fact-checkers may differ in how they frame claims, interpret evidence, or articulate justifications, which can introduce variability not reflective of ground-truth facts but rather of editorial choices. Additionally, reliance on a single aggregation tool like the Fact Check Explorer may result in an incomplete or skewed view of the global fact-checking landscape, under-representing claims from less frequently indexed sources or under-covered topics.

\paragraph{Scope of Evidence}

Our dataset includes only time-series evidence (with textual descriptions), even though real-world claims often require integrating multiple evidence modalities such as reports, tables, charts, or multimedia. While focusing on time series allows us to study a particularly challenging and underexplored aspect of fact verification, the benchmark does not fully represent the broader, multi-modal nature of the fact-checking process.

\section*{Ethics Statement}

\paragraph{Data provenance and licensing}

All evidence series in TSVer originate from Our World in Data (OWID), which redistributes underlying statistics from official bodies (e.g., UN, World Bank) under permissive Creative Commons licences (CC-BY 4.0). We preserve the OWID identifiers, metadata, and citations so that the data lineage remains transparent.

\paragraph{Privacy and anonymisation.}

We did not anonymise any portion of \textsc{TSVer}. All claims are extracted from publicly accessible fact-checking articles that already appear on journalistic websites and reference well-known public figures, institutions, or countries. These named entities, and the temporal and geographic details, are integral to the factual content of each statement and therefore necessary for fact verification.

\section*{Acknowledgement}

Marek Strong was supported by The Alan Turing Institute’s Enrichment Scheme. Andreas \mbox{Vlachos} is supported by the ERC grant AVeriTeC (GA 865958) and the DARPA program SciFy.

\bibliography{papers_references,anthology,custom}

\appendix


\section{Experimental Setup}
As part of the annotation pipeline, we used the Llama-3 \citep{dubey2024llama-906} model for inference. Specifically, we employed the 8B-parameter version in 16-bit precision. Inference was performed with a temperature of \num{1.0} using nucleus sampling \citep{holtzman2019curious-6be}, with a top-p value of \num{0.9}.

All annotation scripts and experiments were run on a machine equipped with a single Quadro RTX 8000 GPU (49GB memory) and 64GB of system RAM.

For querying the baseline models, we performed inference using each model’s official API. For our baseline experiments with Llama, we used Amazon Bedrock’s API instead of the local model to support the full 128k token context window. All API calls were made with default settings unless otherwise specified.

Additionally, we used a combination of GPT-4 \cite{openai2023gpt-4-0bf} and Claude \cite{anthropic2025claude37sonnet} to assist with parts of the codebase. These models were used as general-purpose coding assistants.

\section{Annotation details}

We carried out our annotations with the help of Prolific (\url{https://www.prolific.com/}), an online platform which connects researchers with real people willing to participate in studies and surveys, enabling fast collection of high-quality data. The annotations took place on a separate dedicated platform developed by our team and supplied to Prolific.

To ensure high-quality annotations, we applied participant screening criteria available through Prolific. In particular, we restricted participation to individuals located in the United States whose primary language was English and who had completed at least an undergraduate degree (BA/BSc/other). Participants were compensated at an average rate of £10 per hour, in accordance with Prolific’s payment principles (\url{https://researcher-help.prolific.com/en/article/2273bd}).

\onecolumn

\begin{figure}[H]
\centering
\includegraphics[width=1.0\linewidth, trim={0.5cm 0.1cm 0.5cm 0.1cm},clip]{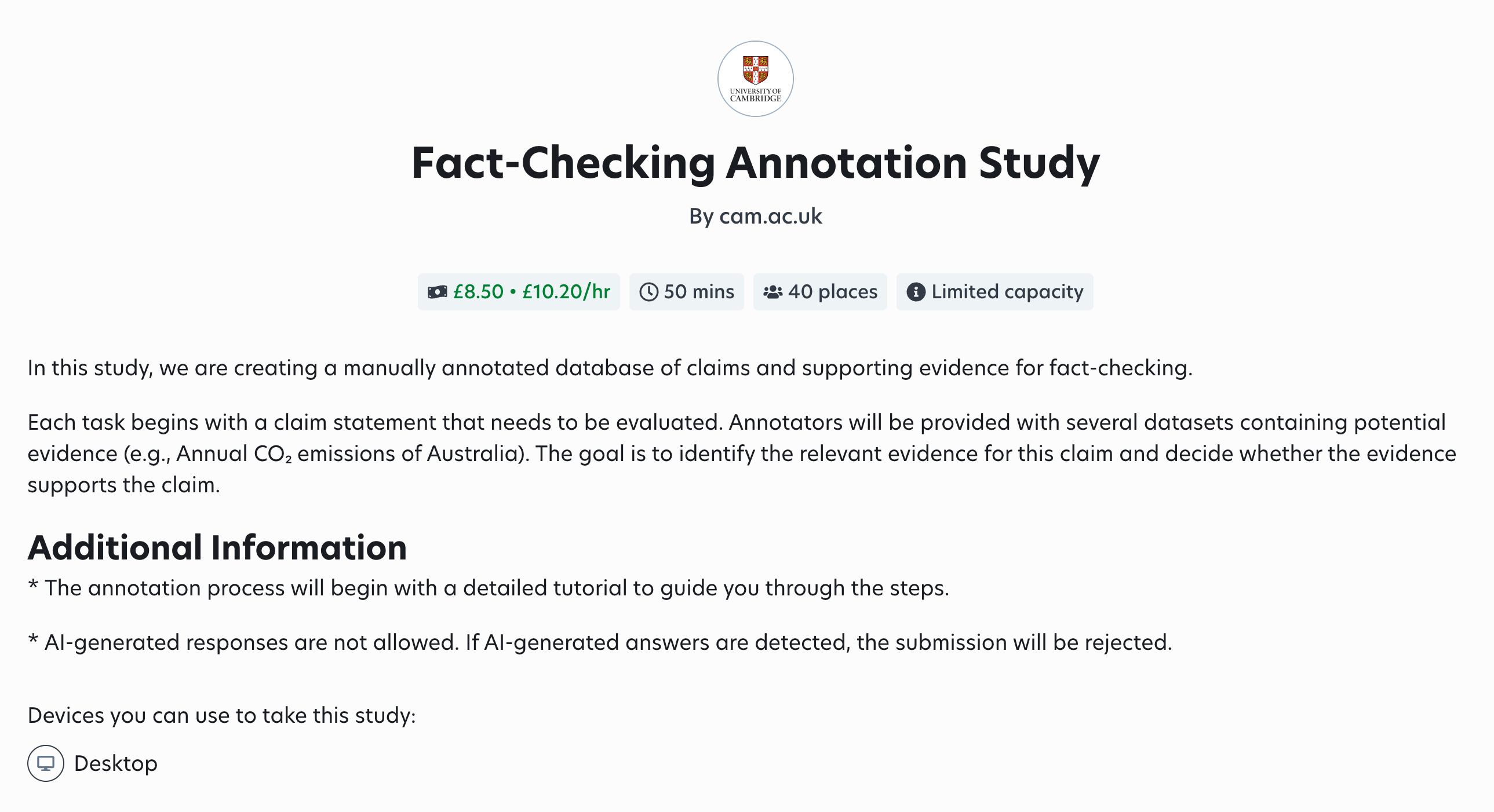}
\caption{Instructions given to participants at the beginning of the annotation session. These instructions were followed by a detailed tutorial.}
\label{fig:prolific-intro}
\end{figure}

\section{Annotation Interface}
\label{sec:appendix_interface}

\begin{figure}[H]
\centering
\includegraphics[width=1.0\linewidth,clip]{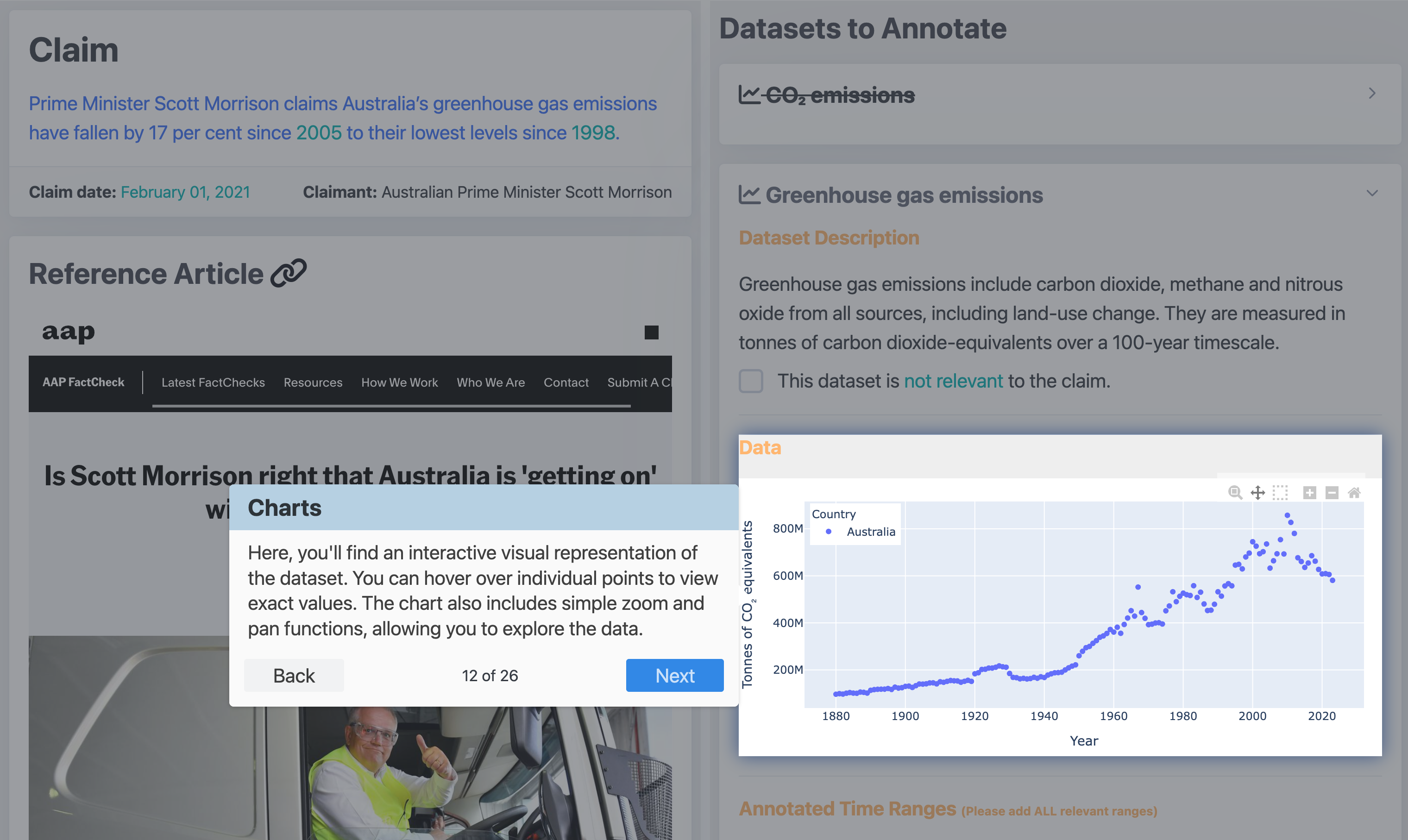}
\caption{\textbf{Detailed step-by-step tutorial explaining the annotation study.}}
\label{fig:portalexample1}
\end{figure}

\begin{figure}[H]
\centering
\includegraphics[width=1.0\linewidth,clip]{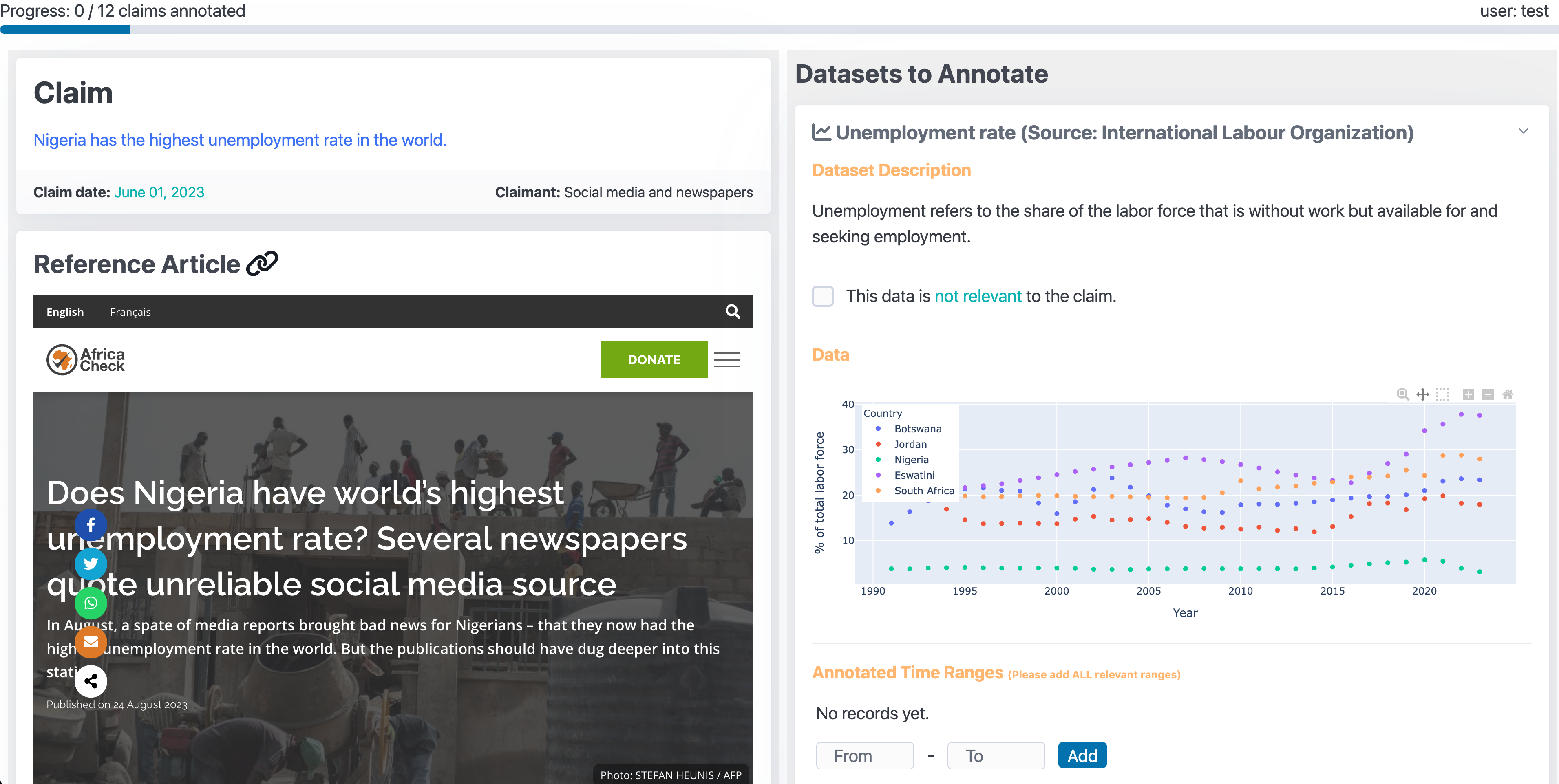}
\caption{\textbf{Annotation Interface for Phase 1.}}
\label{fig:portalexample2}
\end{figure}

\begin{figure}[H]
\centering
\includegraphics[width=1.0\linewidth,clip]{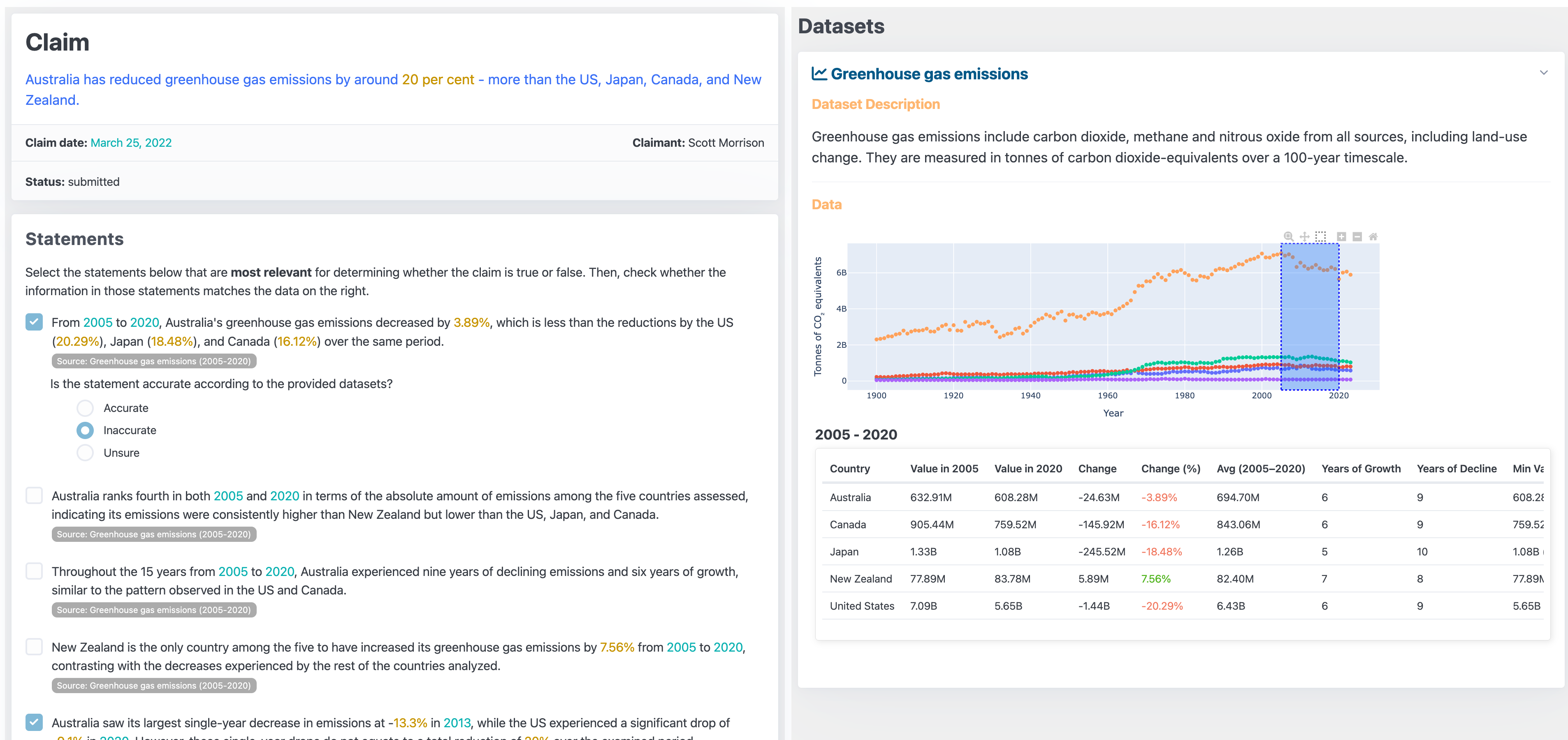}
\caption{\textbf{Annotation Interface for Phase 2.}}
\label{fig:portalexample3}
\end{figure}

\section{Dataset Details}
\label{sec:appendix_datasetdetails}

\begin{figure}[H]
\centering
\includegraphics[width=0.9\linewidth, trim={0.2cm 0.1cm 0.5cm 0.1cm},clip]{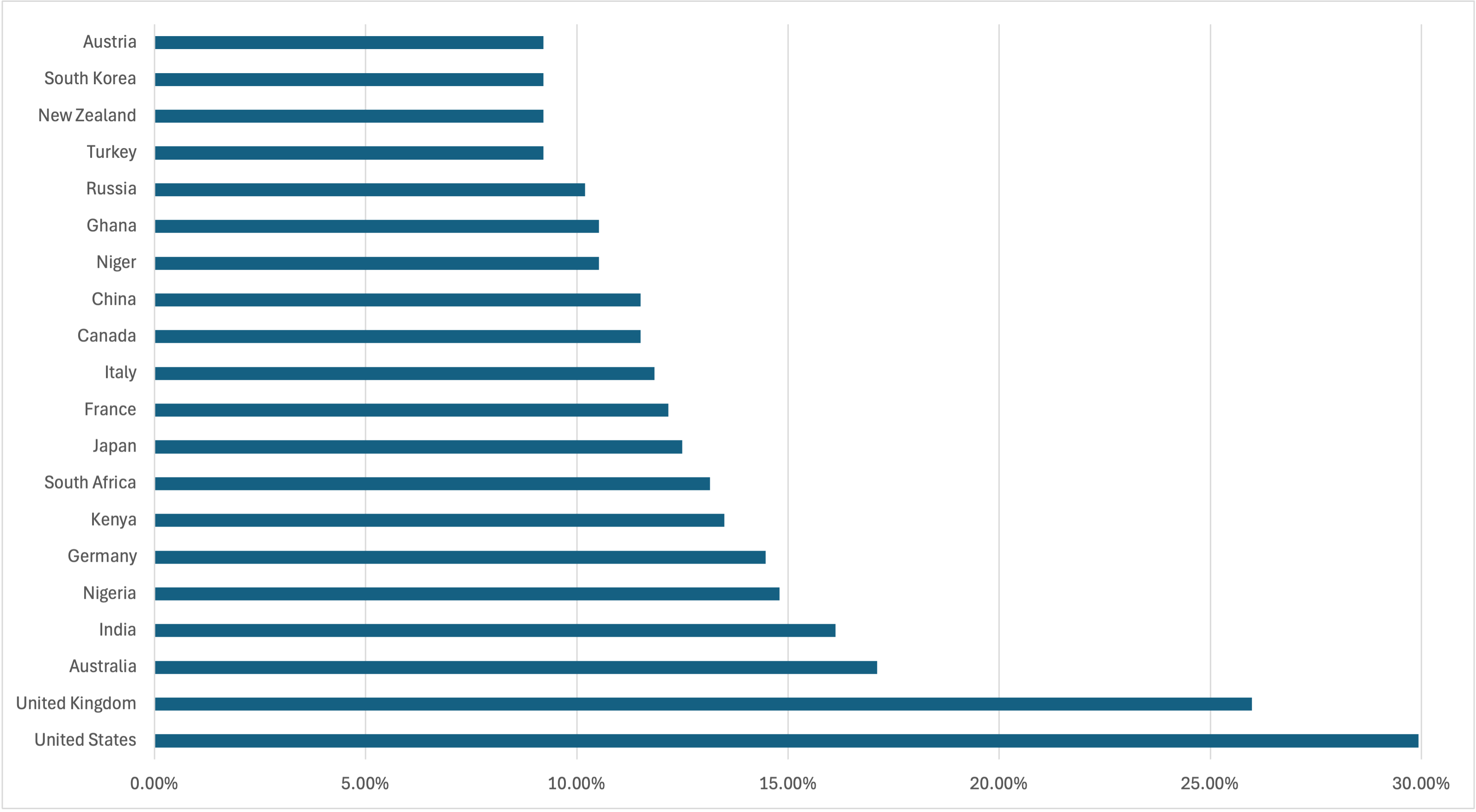}
\caption{Top 20 countries by share of claims in the benchmark dataset. Bars indicate the percentage of claims associated with each country.}
\label{fig:claims-by-country}
\end{figure}

\begin{figure}[H]
\centering
\includegraphics[width=0.9\linewidth, trim={0.5cm 0.1cm 0.5cm 0.1cm},clip]{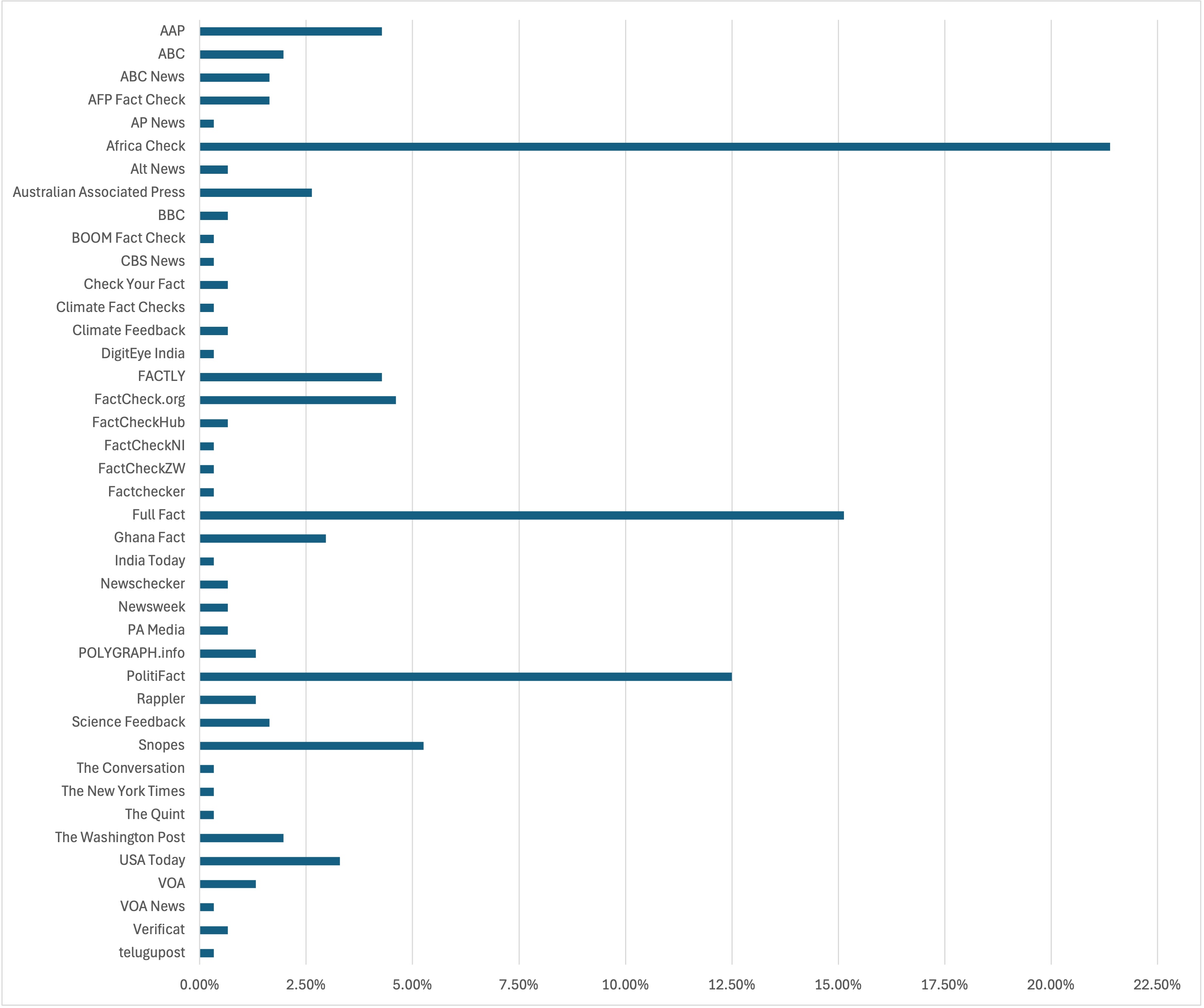}
\caption{Share of fact-checked claims by publishing organization (\emph{N}=\textit{41}) in \emph{TSVer}. Africa Check (21.38\%) accounts for the largest share, followed by Full Fact (15.13\%) and PolitiFact (12.50\%).}
\label{fig:claims-by-org}
\end{figure}

\begin{figure}[H]
\noindent
\begin{lstlisting}[caption={An example of metadata associated with time-series in the TSVer dataset.}, label={appendix:metadata}]

Name: Population growth rate with migration
Description: "Average exponential rate of growth of the population over a given period. It is calculated as ln(P2/P1) where P1 and P2 are the populations on 1 January of subsequent years."
Unit: "%"


Name: Number of people living in urban areas
Description: "Urban population refers to people living in urban areas as defined by national statistical offices. It is calculated using World Bank population estimates and urban ratios from the United Nations World Urbanization Prospects.<br><br>Limitations and exceptions: Aggregation of urban and rural population may not add up to total population because of different country coverage. There is no consistent and universally accepted standard for distinguishing urban from rural areas, in part because of the wide variety of situations across countries.<br><br>Because the estimates of city and metropolitan area are based on national definitions of what constitutes a city or metropolitan area, cross-country comparisons should be made with caution."
Unit: "People"


Name: Population by age group (15-19 years)
Description: "De facto total population in a country, area or region as of 1 July of the year indicated. This only includes individuals aged 15-19."
Unit: "Persons"


Name: Press Freedom Index
Description: "The index combines expert estimates with data on violence against journalists. It ranges from 0 (freedom) to 100 (no freedom). The variable denotes a country's press freedom score. It combines data on violence against journalists with experts assessments by media professionals, lawyers, and sociologists on pluralism, media independence, media environment and self-censorship, legislative framework, transparency, and the quality of the infrastructure that supports the production of news and information. Note: Good = 0-15 points; Satisfactory = 15-25; Problematic = 25-35; Difficult = 35-55; Very serious = 55-100."
Unit: "Index"


Name: Freedom of Expression Score
Description: "Data by the Bertelsmann Transformation Index. Expert estimates of the extent to which individuals, groups, and the press can express their views free from government interference. Scores range from 1 to 10 (most free). Score of 1: Freedom of expression is denied. Independent media do not exist or are prohibited. Score of 4: Freedom of expression is often subject to interference or government restrictions. Distortion and manipulation shape matters of public debate. Score of 7: Freedom of expression is occasionally subject to interference or government restrictions, but there are generally no incidents of blatant intrusions like outright state censorship or media shutdowns. Score of 10: Freedom of expression is guaranteed against interference or government restrictions. Individuals, groups and the press can fully exercise these rights. The remaining scores are intermediate categories."
Unit: "Score"


Name: Annual CO2 emissions from coal
Description: "Annual emissions of carbon dioxide (CO2) from coal, measured in tonnes. This data is based on territorial emissions, meaning the emissions produced within a country's borders, but not those from imported goods. For example, emissions from imported steel are counted in the country where the steel is produced. Emissions from international aviation and shipping are not included in the data for any individual country or region. They are only counted in the global total."
Unit: "tonnes"


Name: Share of the population that is female
Description: "Female population is the percentage of the population that is female. Population is based on the de facto definition of population, which counts all residents regardless of legal status or citizenship."
Unit: "% of total"


Name: Area of planted forest
Description: "Forest predominantly composed of trees established through planting and/or deliberate seeding.Explanatory notes1. In this context, predominantly means that the planted/seeded trees are expected to constitute more than 50 percent of the growing stock at maturity. 2. Includes coppice from trees that were originally planted or seeded."
Unit: "hectares"
\end{lstlisting}
\end{figure}

\section{Prompting}
\label{sec:appendix-prompting}

\begin{figure}[H]
\noindent
\begin{lstlisting}[caption={The prompt template used for statement generation during the second phase of human annotation.}, label={prompt_annotation_statements}]
# Datasets Provided

You have access to the following dataset(s):

{time_series_metadata}

------------------------

# Data

The following section contains the actual data for each dataset.
Each dataset may include one or more time ranges - when multiple time ranges are available, they are provided separately.

{time_series_data}

------------------------

# Claim

Evaluate the following claim, stated on {claim_date}:

{claim_text}

------------------------

# Instructions for Analysis

Based only on the data provided above, generate up to five concise assertions that either support or refute the claim.


Each assertion must follow these guidelines:

1. Data-Backed: Every assertion must cite specific figures, statistics, or rankings from the data. Mention relevant time ranges within the assertion itself. Include the dataset name(s) in brackets at the end of each assertion for attribution, such as [Coal production (2007-2017), Energy consumption (1930-2010), Energy consumption (2003-2004)].

2. Factual Only: Do not include assumptions, interpretations, or projections beyond what the data shows.

3. Data Relevance: Consider whether the data is relevant to when the claim was stated.

4. Clear & Focused: Keep assertions concise and directly tied to the claim.

5. Contextual Language: Refer to data using natural phrasing based on the time periods and content, rather than naming the dataset titles or headings explicitly.

6. Output Format: Use a numbered list (1-5) for your assertions.
\end{lstlisting}
\end{figure}

\begin{figure}[H]
\noindent
\begin{lstlisting}[caption={Prompt template for time series retrieval.}, label={prompt_retriever_tseries}]

# AVAILABLE TIME-SERIES DATA:

{time_series_metadata}

------------------------

# INSTRUCTIONS:

Your task is to identify and list all time-series charts that would be relevant or helpful for verifying a provided claim.

Output your response as a numbered list containing only the titles of the relevant time series charts.

Output the list of relevant time series and say nothing else.

------------------------

# Examples:

{few_shot_examples}

------------------------

# CLAIM:

"{claim_text}"
\end{lstlisting}
\end{figure}

\begin{figure}[H]
\noindent
\begin{lstlisting}[caption={Prompt template for the retrieval of relevant countries.}, label={prompt_retriever_countries}]
# TIME-SERIES DATA:

{time_series_metadata}


# AVAILABLE COUNTRIES:

{country_names}

------------------------

# INSTRUCTIONS:

Your task is to identify and list all countries that would be relevant or helpful for verifying a provided claim.

When verifying the claim, we will also have access to time series evidence data as listed above.

Provide your response as a numbered list containing only countries provided in the list above.

Output the list of relevant countries and say nothing else.


------------------------

# Examples:

{few_shot_examples}

------------------------

# CLAIM:

"{claim_text}"
\end{lstlisting}
\end{figure}

\begin{figure}[H]
\noindent
\begin{lstlisting}[caption={Prompt template for the retrieval of relevant time ranges.}, label={prompt_retriever_tranges}]
# TIME SERIES DATA:

{time_series_metadata}


# CLAIM DATE:

{claim_date}


------------------------


# INSTRUCTIONS:

Your task is to identify all time ranges in the provided time series metadata that could be relevant or helpful for verifying the claim made on {claim_date}.

Output your response as a bullet-point list for each time series, using the following format:

# Time-Series-Name-1
- YYYY-YYYY
- YYYY

# Time-Series-Name-2
- YYYY

# Time-Series-Name-3
- YYYY-YYYY
- YYYY-YYYY
- YYYY-YYYY


Output only the list of relevant time ranges for each time series. Do not include any additional text.


------------------------

# Examples:

{few_shot_examples}

------------------------

# CLAIM:

"{claim_text}"
\end{lstlisting}
\end{figure}

\begin{figure}[H]
\noindent
\begin{lstlisting}[caption={Prompt template for verdict and justification generation without CoT.}, label={prompt_verdict_2step}]
Consider the following claim, stated on {claim_date} by {claimant}:
"{claim_text}"

------------------------

Your task is to assess the veracity of this claim using the provided time series data.

# TIME SERIES DATA:

{relevant_tseries_data}

------------------------

# INSTRUCTIONS:
Evaluate the claim in two steps:
- First, select a verdict based on the time series data.
- Second, provide a brief explanation justifying your verdict.

When choosing the verdict, you can choose only from the following options:
{labels_legend}

Format your response as follows:
# VERDICT
...

# EXPLANATION
...
\end{lstlisting}
\end{figure}

\begin{figure}[H]
\noindent
\begin{lstlisting}[caption={Prompt template for verdict and justification generation with CoT.}, label={prompt_verdict_3step}]
Consider the following claim, stated on {claim_date} by {claimant}:
"{claim_text}"

------------------------

Your task is to assess the veracity of this claim using the provided time series data.

# TIME SERIES DATA:

{relevant_tseries_data}

------------------------

# INSTRUCTIONS:
Evaluate the claim in three steps:
- First, reason about the data step by step.
- Second, select a verdict based on the time series data.
- Third, provide a brief explanation justifying your verdict.

When choosing the verdict, you can choose only from the following options:
{labels_legend}

Format your response as follows:
# REASONING
...

# VERDICT
...

# EXPLANATION
...
\end{lstlisting}
\end{figure}

\begin{figure}[H]
\noindent
\begin{lstlisting}[caption={Prompt template for the Ev2R scorer.}, label={prompt_ev2r}]
You will get as input a claim, a reference evidence and a predicted evidence.
Please verify the correctness of the predicted evidence by comparing it to the reference evidence, following these steps:

1. Break down the PREDICTED evidence in independent facts. Each fact should be a separate sentence.
2. Evaluate each fact individually: is the fact supported by the REFERENCE evidence? Do not use additional sources or background knowledge.
3. Next, break down the REFERENCE evidence in independent facts. Each fact should be a separate sentence.
4. Evaluate each fact individually: is the fact supported by the PREDICTED evidence? Do not use additional sources or background knowledge.
5. Finally summarise (1.) how many predicted facts are supported by the reference evidence, (2.) how many reference facts are supported by the predicted evidence.

Generate the output in form of a JSON as shown in the examples below.

----------

Examples:

{few_shot_examples}

----------

Input:

# CLAIM:
"{claim_text}"

# REFERENCE EVIDENCE:
"{reference_evidence}"

# PREDICTED EVIDENCE:
"{predicted_evidence}"
\end{lstlisting}
\end{figure}

\section{Annotation Details}
\label{sec:appendix_annotation_details}

\begin{table}[H]
    \centering
    \resizebox{0.95\linewidth}{!}{
    \begin{tabular}{@{}lll@{}}
    \toprule
    \textbf{Name of Operation}                         & \textbf{Description}                                                                             & \textbf{Example Output}               \\ \midrule
    \multicolumn{1}{|l|}{Difference (Change)}          & \multicolumn{1}{l|}{Computes the absolute change in value between the start and end year.}       & \multicolumn{1}{l|}{150.0}            \\ \midrule
    \multicolumn{1}{|l|}{Percent Change}               & \multicolumn{1}{l|}{Computes the percentage change from the starting value to the ending value.} & \multicolumn{1}{l|}{12.5\%}           \\ \midrule
    \multicolumn{1}{|l|}{Average}                      & \multicolumn{1}{l|}{Calculates the mean value across the selected years.}                        & \multicolumn{1}{l|}{123.45}           \\ \midrule
    \multicolumn{1}{|l|}{Cumulative Total}             & \multicolumn{1}{l|}{Sums all values over the selected time range.}                               & \multicolumn{1}{l|}{987.65}           \\ \midrule
    \multicolumn{1}{|l|}{Standard Deviation}           & \multicolumn{1}{l|}{Measures variability or dispersion from the average value.}                  & \multicolumn{1}{l|}{15.23}            \\ \midrule
    \multicolumn{1}{|l|}{Minimum Value}                & \multicolumn{1}{l|}{Finds the lowest value in the time range.}                                   & \multicolumn{1}{l|}{100.00}           \\ \midrule
    \multicolumn{1}{|l|}{Maximum Value}                & \multicolumn{1}{l|}{Finds the highest value in the time range.}                                  & \multicolumn{1}{l|}{200.00}           \\ \midrule
    \multicolumn{1}{|l|}{Number of Years of Growth}    & \multicolumn{1}{l|}{Counts years where the value increased from the previous year.}              & \multicolumn{1}{l|}{4}                \\ \midrule
    \multicolumn{1}{|l|}{Number of Years of Decline}   & \multicolumn{1}{l|}{Counts years where the value decreased from the previous year.}              & \multicolumn{1}{l|}{3}                \\ \midrule
    \multicolumn{1}{|l|}{Largest Single-Year Drop}     & \multicolumn{1}{l|}{Finds the largest decrease in value between two consecutive years.}          & \multicolumn{1}{l|}{-45.67 (in 2012)} \\ \midrule
    \multicolumn{1}{|l|}{Largest Single-Year Increase} & \multicolumn{1}{l|}{Finds the largest increase in value between two consecutive years.}          & \multicolumn{1}{l|}{55.23 (in 2018)}  \\ \midrule
    \multicolumn{1}{|l|}{Average Rank}                 & \multicolumn{1}{l|}{Computes the average ranking of a country over the selected years.}          & \multicolumn{1}{l|}{2.4}              \\ \midrule
    \multicolumn{1}{|l|}{Rank in Year}                 & \multicolumn{1}{l|}{Ranks countries by value in a specific year (1 = highest value). }          & \multicolumn{1}{l|}{1.0}              \\ \midrule
    \end{tabular}
}
\caption{List of operators with descriptions used for generating pre-computed statistics in the second annotation phase.}
    \label{tab:appendix-operators}
\end{table}

\end{document}